\documentclass{article}

\usepackage{tabularx}
\usepackage{makecell}

\usepackage[letterpaper,top=2cm,bottom=2cm,left=3cm,right=3cm,marginparwidth=1.75cm]{geometry}

\usepackage{indentfirst}   
\usepackage{amsmath, amssymb}  
\usepackage{graphicx}
\usepackage{booktabs}
\usepackage{siunitx}
\usepackage{caption}
\usepackage{subcaption}

\usepackage[colorlinks=true, allcolors=blue]{hyperref}
\usepackage{cleveref}            
\usepackage{comment}
\usepackage{color,soul}

\title{AIRoA MoMa Dataset: A Large-Scale Hierarchical Dataset \\for Mobile Manipulation}
\author{
\parbox{0.95\linewidth}{\centering
\small
Ryosuke Takanami$^{1}$$^\ast$, Petr Khrapchenkov$^{1,2}$$^\ast$, Shu Morikuni$^{1}$$^\ast$, Jumpei Arima$^{2,3}$, 
Yuta Takaba$^{2,3}$, Shunsuke Maeda$^{2,3}$, Takuya Okubo$^{1}$, Genki Sano$^{4}$, 
Satoshi Sekioka$^{2}$, Aoi Kadoya$^{2}$, \\
Motonari Kambara$^{1}$, Naoya Nishiura$^{1}$, 
Haruto Suzuki$^{1}$, Takanori Yoshimoto$^{1}$, Koya Sakamoto$^{1}$, Shinnosuke Ono$^{1}$, 
Hu Yang$^{1}$, Daichi Yashima$^{1}$, Aoi Horo$^{1}$, Tomohiro Motoda$^{5}$, 
Kensuke Chiyoma$^{2}$, \\
Hiroshi Ito$^{6}$, Koki Fukuda$^{1}$, Akihito Goto$^{2,3}$, 
Kazumi Morinaga$^{5}$, Yuya Ikeda$^{1}$, Riko Kawada$^{6}$,  \\
Masaki Yoshikawa$^{6}$, 
Norio Kosuge$^{2}$, Yuki Noguchi$^{2,3}$, 
Kei Ota$^{2}$, Tatsuya Matsushima$^{1,2}$, \\
Yusuke Iwasawa$^{1,2}$, Yutaka Matsuo$^{1,2}$, 
Tetsuya Ogata$^{2,6}$ \\
\vspace{2ex}
$^{1}$The University of Tokyo \quad
$^{2}$AI Robot Association (AIRoA) \quad
$^{3}$Toyota Motor Corporation \quad
$^{4}$Telexistence, Inc. 
$^{5}$National Institute of Advanced Industrial Science and Technology (AIST) \quad
$^{6}$Waseda University 
$^\ast$indicates equal contribution
}
}

\date{}

\begin{document}
\maketitle

\begin{figure}[h]
    \centering
    \vspace{-5mm}
    \includegraphics[width=0.85\linewidth]{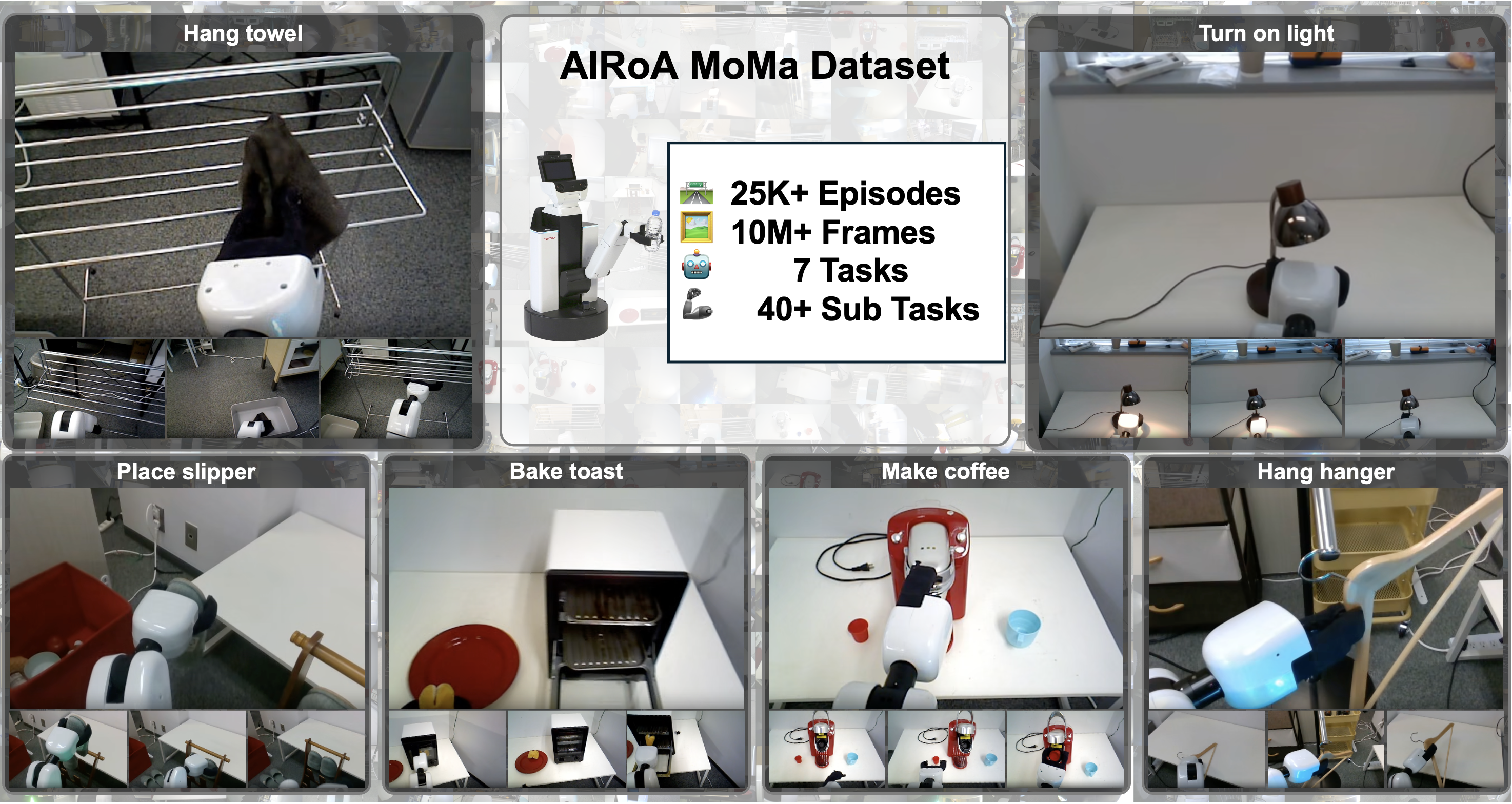}
    \caption{\textbf{Overview of the AIRoA MoMa Dataset.} 
    The dataset comprises over 25,000 episodes with hierarchical annotations covering seven primary tasks and more than 40 sub-tasks. 
    It is designed for mobile manipulation in household settings and provides synchronized multimodal data streams, including force–torque measurements. Representative examples include contact-rich tasks such as``Hang towel,'' ``Turn on light,'' and ``Make coffee.''
    }
    \vspace{-5mm}
    \label{fig:intro_figure}
\end{figure}

\begin{abstract}

As robots transition from controlled settings to unstructured human environments, building generalist agents that can reliably follow natural language instructions remains a central challenge. Progress in robust mobile manipulation requires large-scale multimodal datasets that capture contact-rich and long-horizon tasks, yet existing resources lack synchronized force-torque sensing, hierarchical annotations, and explicit failure cases.
We address this gap with the AIRoA MoMa Dataset, a large-scale real-world multimodal dataset for mobile manipulation. It includes synchronized RGB images, joint states, six-axis wrist force-torque signals, and internal robot states, together with a novel two-layer annotation schema of sub-goals and primitive actions for hierarchical learning and error analysis. The initial dataset comprises 25,469 episodes ($\approx$94 hours) collected with the Human Support Robot (HSR) and is fully standardized in the LeRobot v2.1 format.
By uniquely integrating mobile manipulation, contact-rich interaction, and long-horizon structure, AIRoA MoMa provides a critical benchmark for advancing the next generation of Vision-Language-Action models.
The first version of our dataset is now available in  \href{https://huggingface.co/datasets/airoa-org/airoa-moma}{our dataset repository}.

\end{abstract}

\section{Introduction}

The development of general-purpose robots capable of operating in everyday human environments remains a grand challenge in robotics.
Vision-Language-Action (VLA) models have recently emerged as a promising paradigm toward this goal, linking natural language instructions with real-world robotic actions. However, the performance and generalization of these models are fundamentally constrained by the scale, diversity, and quality of the datasets on which they are trained. While large-scale aggregation efforts such as Open X-Embodiment~\cite{oxe2023} have demonstrated the value of data-driven approaches, existing resources fall short of capturing the multi-step, physically interactive (a.k.a. contact-rich) nature of real-world household tasks.

Specifically, current datasets suffer from three major limitations. First, they focus predominantly on fixed-base tabletop manipulation, leaving mobile manipulation---where navigation and manipulation must be seamlessly integrated---largely unexplored. Second, they rarely capture contact-rich interactions (e.g., pressing switches, opening doors), which demand multimodal sensing such as force–torque feedback beyond vision alone. Third, they underrepresent long-horizon trajectories, where complex tasks (e.g., “make toast”) require hierarchical decomposition into sub-goals and primitive actions. These gaps represent not only underexplored areas but also fundamental barriers preventing the community from advancing beyond short-horizon pick-and-place scenarios toward robust, real-world assistance.

To directly address this bottleneck, we introduce the \emph{AIRoA MoMa Dataset}, a large-scale, multimodal, real-robot dataset explicitly designed to address mobile manipulation, contact-rich interaction, and long-horizon tasks. Our dataset integrates synchronized 6-axis wrist force–torque signals with visual and proprioceptive data, enabling learning of physically grounded interactions. We further introduce a novel two-layer annotation scheme of sub-goals and primitive actions, which facilitates hierarchical learning and fine-grained error analysis. Finally, to ensure portability and reproducibility, we release an open-source pipeline that standardizes the dataset into the widely adopted LeRobot format.

We publicly release the dataset in a format immediately compatible with existing VLA architectures, providing both a benchmark and a resource to accelerate community progress towards the next generation of general-purpose robots. The main contributions of this paper are as follows:

\begin{enumerate}
    \item A large-scale multimodal dataset that uniquely integrates mobile manipulation, contact-rich interaction, and long-horizon tasks.
    \item A hierarchical two-layer annotation schema of sub-goals and primitive actions that facilitates structured learning and detailed failure analysis.
    \item Inclusion of explicit failure cases to support research on error detection and recovery.
    \item An open-source pipeline that standardizes data into the LeRobot format, ensuring reproducibility and broad accessibility.
\end{enumerate}

\section{Related Work}

\subsection{Robot Foundation Models for Mobile Manipulator Robots}

A long-standing ambition in Artificial Intelligence has been to develop general-purpose robots capable of seamlessly operating in diverse environments, manipulating arbitrary objects, and flexibly applying a wide range of skills to accomplish varied tasks. 
Specifically for a mobile manipulation platform, one notable milestone is RT-1 by Google \cite{brohan2023_rt1}. RT-1 explored whether robotic models can benefit from large-scale information drawn from different domains and whether they can exhibit effective zero-shot generalization to novel tasks, environments, and objects.
Built on a transformer architecture, RT-1 processes a history window of images along with natural language instructions to directly output tokenized actions. The model demonstrated the ability to effectively leverage heterogeneous data sources, including simulations and diverse robotic embodiments. RT-1 achieved an impressive 97\% success rate on its trained tasks. More importantly, it generalized learned behaviors to entirely unseen tasks with a 76\% success rate, maintained robustness in cluttered environments at 83\%, and adapted to unfamiliar settings with 59\% success.
Its successor, RT-2~\cite{zitkovich2023_rt2}, extended these capabilities by incorporating web-scale data. RT-2 demonstrated stronger generalization, with enhanced semantic and visual scene understanding that went beyond the original robotic training data. This included the ability to interpret novel natural language instructions and to perform higher-level reasoning about object descriptions.

Another prominent area of research, particularly concerning mobile manipulator platforms, involves the Mobile ALOHA system~\cite{fu2024mobile}.
Developed by researchers at Stanford University, Mobile ALOHA integrates mobility and bimanual manipulation capabilities to autonomously execute complex household tasks. Leveraging a learning by imitation framework, the robot is trained by observing and precisely reproducing human demonstrations. This methodology enables Mobile ALOHA to demonstrate proficiency across a diverse array of tasks, achieving an efficiency that motivates a rethink collaboration between humans and robots.

While RT-1 for example showed generalization across instructions and various robot embodiments, RT-2 successfully transferred knowledge from web-scale sources, and Mobile ALOHA provides cost-effective solution for a whole-body teleoperation interface, several challenges still remain.
(1) The model primarily relies on positional information for robot proprioception, lacking crucial force feedback.
(2) most of the mobile manipulator uses discrete action.
(3) As the authors themselves acknowledge, although incorporating web-scale pretraining through VLMs enhances generalization over semantic and visual concepts, it does not endow the robot with the ability to acquire new motion skills from this additional experience. To achieve human-level competency will require significantly more diverse robot data spanning a wide range of objects, environments, tasks, and situations.

\subsection{Large-Scale Real-World Robot Datasets}

\begin{table}[t]
  \centering
  \caption{Comparison of large-scale real-world robot datasets. The definition of Tasks/Scenes varies across studies (e.g., verb-based). OXE is an integrated repository across multiple research institutions. 
  }
  \setlength{\tabcolsep}{4pt} %
  \footnotesize %
  \begin{tabularx}{\textwidth}{l c c c X X >{\raggedright\arraybackslash}X} %
    \toprule
    \thead{Dataset \\ (Year)} & \thead{Traj./ \\ Hours} & \thead{Tasks} & 
    \thead{Scenes} & \thead{Embodiments} & \thead{Env.} & 
    \thead{Unique Modals} \\
    \midrule
    RT-1 \cite{brohan2023_rt1} & $\sim$130k / n.a. & 700 & n.a. & 13 robots & almost tabletop & - \\
    OXE / RT-X \cite{oxe2023} & $>1$M / n.a. & 527 skills & $\sim$311 & 22 robots & alomost tabletop & - \\
    DROID \cite{khazatsky2024_droid} & 76k / 350h & 86 & 564 & 1 (Franka) & tabletop & Depth, Calib \\
    BridgeData V2 \cite{walke2023_bridgedata} & 60{,}096 / n.a. & 13 skills & 24 & 1 (WidowX 250) & tabletop & - \\
    RH20T \cite{fang2024_rh20t} & $>110$k / n.a. & $>140$ & 7 & 4 robots & tabletop & Force, Audio \\
    RoboNet \cite{dasari2019robonet} & $\sim$162k / 15M frames & n.a. & n.a. & 7 robots & tabletop & - \\
    \bottomrule
  \end{tabularx}
  \label{tab:real_robot_datasets}
\end{table}

The foundation for the advancement of VLA models is large-scale, highly diverse real-world datasets. Key examples are summarized in Table~\ref{tab:real_robot_datasets}. RT-1 demonstrated generalization to over 700 instructions using \SI{130}{k} episodes collected from 13 robots over 17 months, and RT-2 demonstrated the transfer of web-scale knowledge. The OXE/RT-X~\cite{oxe2023} dataset integrated over one million trajectories from 22 different robot embodiments, showing positive transfer in cross-embodiment learning. DROID contains \SI{76}{k} trajectories (\SI{350}{h}) across 86 tasks and 564 scenes, including multi-view RGB/Depth data, language annotations, and calibration information~\cite{khazatsky2024_droid}. BridgeData V2 consists of \SI{60096}{} trajectories for 82 tasks in 24 environments, conditioned on goal images and language~\cite{walke2023_bridgedata}. DROID dataset has taken steps toward increasing scene diversity by crowdsourcing demonstrations. RH20T includes over 110,000 contact-rich, multimodal trajectories featuring vision, force, audio, language, and action data~\cite{fang2024_rh20t}. RoboNet provides approximately \SI{162}{k} trajectories (around \SI{15}{M} frames) from seven different robot platforms~\cite{dasari2019robonet}.

Existing large-scale real-robot datasets, while focused primarily on tabletop manipulation and short-horizon tasks using visual and language modalities, remain highly constrained. Designed for narrow tasks, trained on limited data, and deployed in fixed environments, these datasets result in models that exhibit poor generalization and lack robustness to distribution shifts in real-world settings. Crucially, current data collections have not sufficiently covered essential aspects such as long-horizon tasks, contact-rich interactions, mobile manipulation, and the systematic collection of failure cases. To address these deficits, this study introduces the AIRoA MoMa Dataset. This real-robot multimodal dataset is specifically designed to include these elements, promoting the advancement of Vision-Language-Action (VLA) model research in the field of mobile manipulation.

\section{AIRoA MoMa Dataset}

To enable foundation models that generalize beyond table-top pick-and-place tasks, we construct the \emph{AIRoA MoMa (Mobile Manipulation) Dataset} using mobile manipulators in realistic household environments.
This dataset focuses on contact-rich interactions and multi-step daily activities, providing synchronized multimodal signals and hierarchical annotations. 
This section describes the data collection setup, the teleoperation system, the dataset scale and modalities, the task hierarchy, and data quality considerations.

\subsection{Overview}

\begin{figure}
    \centering
    \includegraphics[width=1\linewidth]{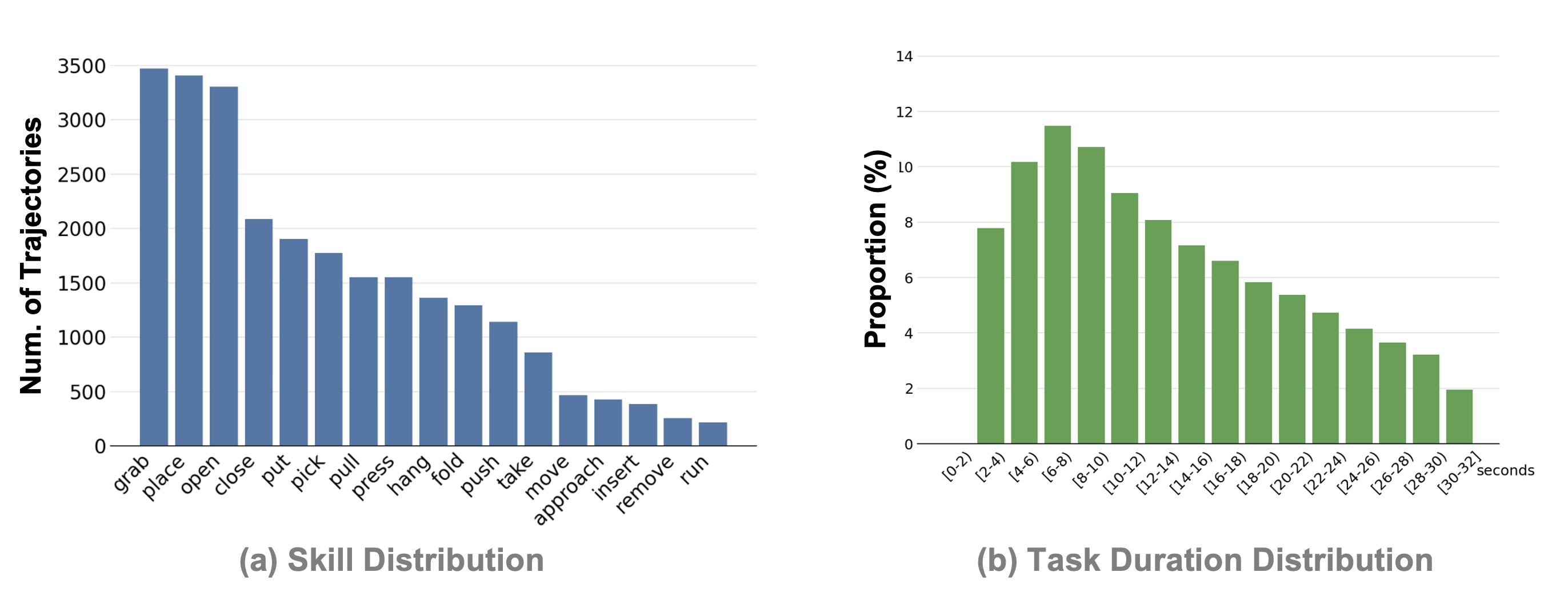}
    \caption{\textbf{Statistics of skill and duration of primitive actions.} \textbf{(a)} The skill distribution exhibits a long-tail pattern, where fundamental manipulation skills such as ``grab,'' ``open,'' and ``place'' constitute the majority of the collected trajectories. \textbf{(b)} The task duration distribution is concentrated in the short-to-medium range, with most tasks being completed in 4 to 12 seconds. Together, these characteristics show that the dataset is primarily composed of discrete, short-horizon activities , making it an ideal resource for training the foundational and reactive policies required for hierarchical approaches to long-horizon tasks.}
    \label{fig:dataset_statistics}
\end{figure}

\textbf{Environments.}
The data collection was conducted in a laboratory replicating household settings, including kitchen, living room, and bathroom as shown in Fig.~\ref{fig:intro_figure}. 
To enhance data diversity, object placement, lighting conditions, and the robot's initial position were randomized across episodes.

\textbf{Modalities}
The dataset provides rich, time-synchronized, multi-modal data streams resampled to 30 Hz. Each timestep includes:
\begin{itemize}
    \item Visual: RGB images (480 x 640 x 3) from two viewpoints: a head-mounted camera (global view) and a wrist-mounted camera (close-up for manipulation).
    \item Proprioceptive: Joint angles of arm, gripper, and lifting torso, as well as velocity of robot base and end-effector pose.
    \item Force-Torque: Six-axis wrist force-torque signals (Fx, Fy, Fz, Mx, My, Mz) essential for contact-rich tasks.
    \item Teleoperation Signals: Raw operator control commands from the teleoperation system preserved for potential use in action representation studies.
\end{itemize}

\textbf{Dataset Scale.}
The dataset consists of 25,469 episodes, totaling approximately 94 hours, with an average episode duration of 13 seconds (see Table.~\ref{tab:episode-summary}). The total dataset size is approximately 92 GB. Seven representative household tasks were included, covering both placement and fine-grained contact manipulation.

\textbf{Action Space.}
We define the robot joint state as the concatenation of arm and wrist joints (5 DoF), gripper (1 DoF), and head joints (2 DoF) encoder observations:
\[
s = \{s^{\text{hsr}}_{\text{arm\_lift}}, s^{\text{hsr}}_{\text{arm\_flex}}, s^{\text{hsr}}_{\text{arm\_roll}}, 
      s^{\text{hsr}}_{\text{wrist\_flex}}, s^{\text{hsr}}_{\text{wrist\_roll}}, 
      s^{\text{hsr}}_{\text{hand\_motor}}, 
      s^{\text{hsr}}_{\text{head\_pan}}, s^{\text{hsr}}_{\text{head\_tilt}}\} \in \mathbb{R}^8.
\]

Absolute actions are defined as joint commands recorded from the teleoperation device, in the same order as the state vector:
\[
a_{\text{absolute}} = \{s^{\text{teleop}}_{\text{arm\_lift}}, s^{\text{teleop}}_{\text{arm\_flex}}, s^{\text{teleop}}_{\text{arm\_roll}}, 
      s^{\text{teleop}}_{\text{wrist\_flex}}, s^{\text{teleop}}_{\text{wrist\_roll}}, 
      s^{\text{teleop}}_{\text{hand\_motor}}, 
      s^{\text{teleop}}_{\text{head\_pan}}, s^{\text{teleop}}_{\text{head\_tilt}}\} \in \mathbb{R}^8.
\]

For more granular control, we also provide action vectors per robot subsystem:
\[
\begin{aligned}
a_{\text{arm}}     &= \{s^{\text{teleop}}_{\text{arm\_lift}}, s^{\text{teleop}}_{\text{arm\_flex}}, s^{\text{teleop}}_{\text{arm\_roll}}, s^{\text{teleop}}_{\text{wrist\_flex}}, s^{\text{teleop}}_{\text{wrist\_roll}}\} \in \mathbb{R}^5, \\
a_{\text{gripper}} &= \{s^{\text{teleop}}_{\text{hand\_motor}}\} \in \mathbb{R}^1, \\
a_{\text{head}}    &= \{s^{\text{teleop}}_{\text{head\_pan}}, s^{\text{teleop}}_{\text{head\_tilt}}\} \in \mathbb{R}^2, \\
a_{\text{base}}    &= \{\Delta x, \Delta y, \Delta \theta\} \in \mathbb{R}^3.
\end{aligned}
\]

All joint actions are provided as absolute commands, obtained from the teleoperation device, with the exception of base motion, which is represented as relative increments $(\Delta x, \Delta y, \Delta \theta)$ derived from interface commands. We do not include absolute base state or action due to the limited accuracy of robot odometry.

Finally, we provide a relative action representation, where arm and head actions are expressed as deltas with respect to the current robot state, the gripper remains absolute, and base motion increments are appended:
\[
a_{\text{relative}} = \{a_{\text{arm}} - s^{\text{hsr}}_{\text{arm}},\; 
                       s^{\text{teleop}}_{\text{gripper}},\; 
                       a_{\text{head}} - s^{\text{hsr}}_{\text{head}},\; 
                       a_{\text{base}}\} \in \mathbb{R}^{11}.
\]

\subsection{Data Collection}

\begin{figure}
    \centering
    \includegraphics[width=1\linewidth]{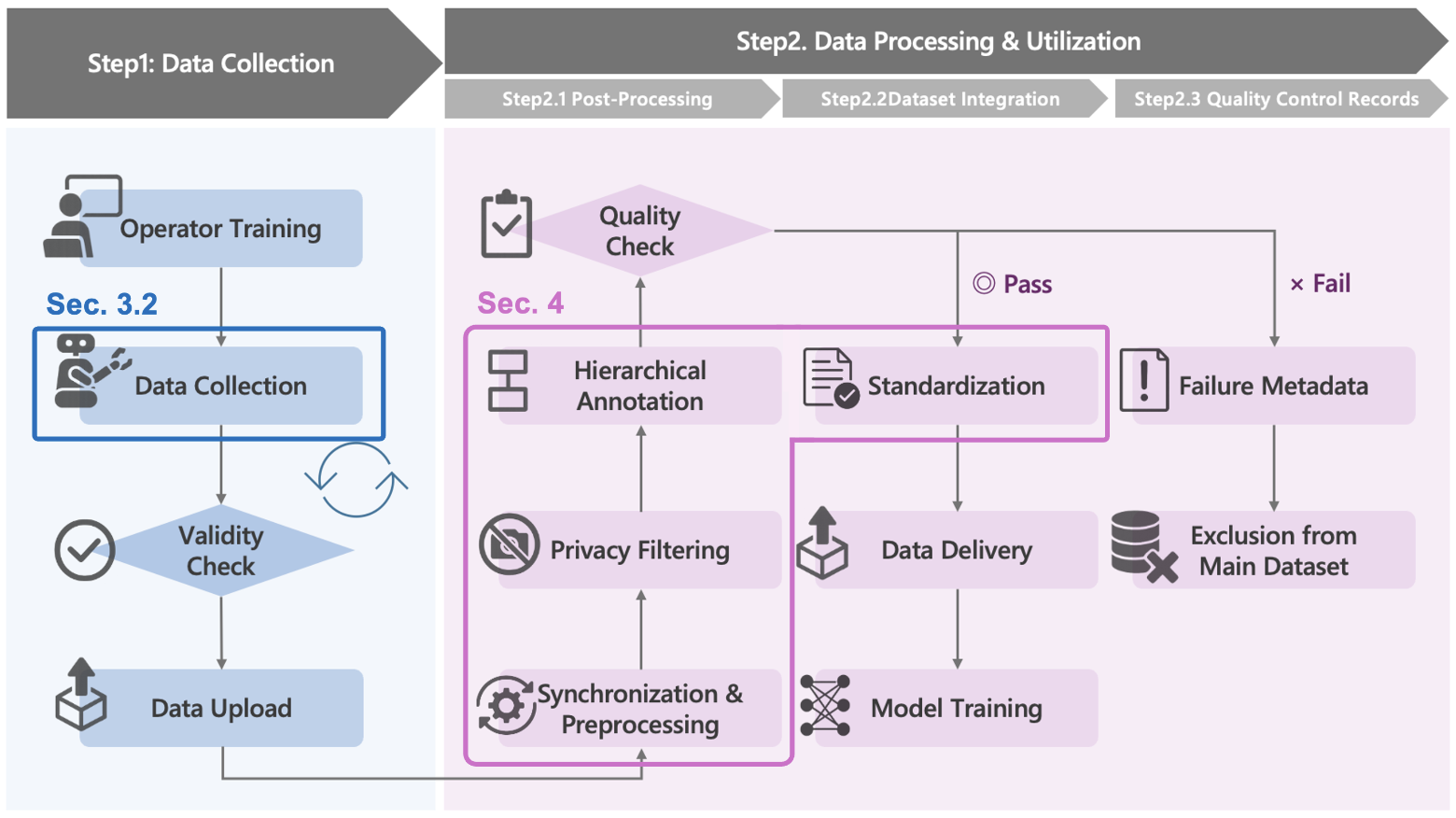}
    \caption{\textbf{The systematic pipeline from data collection to model training.} The pipeline begins with data collection by trained operators, followed by an initial validity check. Uploaded data undergoes a rigorous quality check, after which it is processed through key stages including the core hierarchical annotation, privacy filtering, and synchronization. Finally, the data is standardized into the LeRobot format, ensuring it is delivered in a state ready for immediate use in model training by the research community.}
    \label{fig:pipe}
\end{figure}

\subsubsection{Robot Platform}
We employ the Toyota Human Support Robot (HSR)~\cite{yamamoto2019_hsr}, a versatile personal assistant robot equipped with a 4-DOF manipulator on a lifting torso, 1-DoF hand, 2-DoF head, and 3-DoF omnidirectional mobile base, as our data collection platform (see Fig. \ref{fig:leader_follower_system}). The HSR has been widely adopted by more than 60 research institutions worldwide and serves as the standard platform for the RoboCup@Home league \cite{robocupathome}.

\subsubsection{Teleoperation System}
While robotic pick-and-place tasks have become highly advanced in recent years \cite{matsushima2022_robocup}, the automation of diverse behaviors involving environmental interaction, such as opening drawers or plugging and unplugging cords, remains underdeveloped. Human teleoperation is an effective approach for efficiently collecting such complex behavioral data \cite{philipp2024_gello}\cite{fu2024mobile}. However, for the teleoperation of the HSR the default option is a gamepad controller, which operates on a single-axis basis and is not well-suited for complex maneuvers. Furthermore, methods that command the robot in the end-effector space using devices like VR controllers~\cite{philipp2024_gello, jeffrey2018_baxter, varad2024_mice} which poses it's own challenges, as inverse kinematics (IK) calculations are complicated by the need to include the lifting torso and mobile base, as the HSR arm has only 4-DOF \cite{nakanishi2020_vr}. This approach presents operability and safety concerns, as it can lead to unintended robot postures or collisions between the mobile base and the environment.

\begin{figure}[t]
  \centering
  \includegraphics[width=1.0\linewidth]{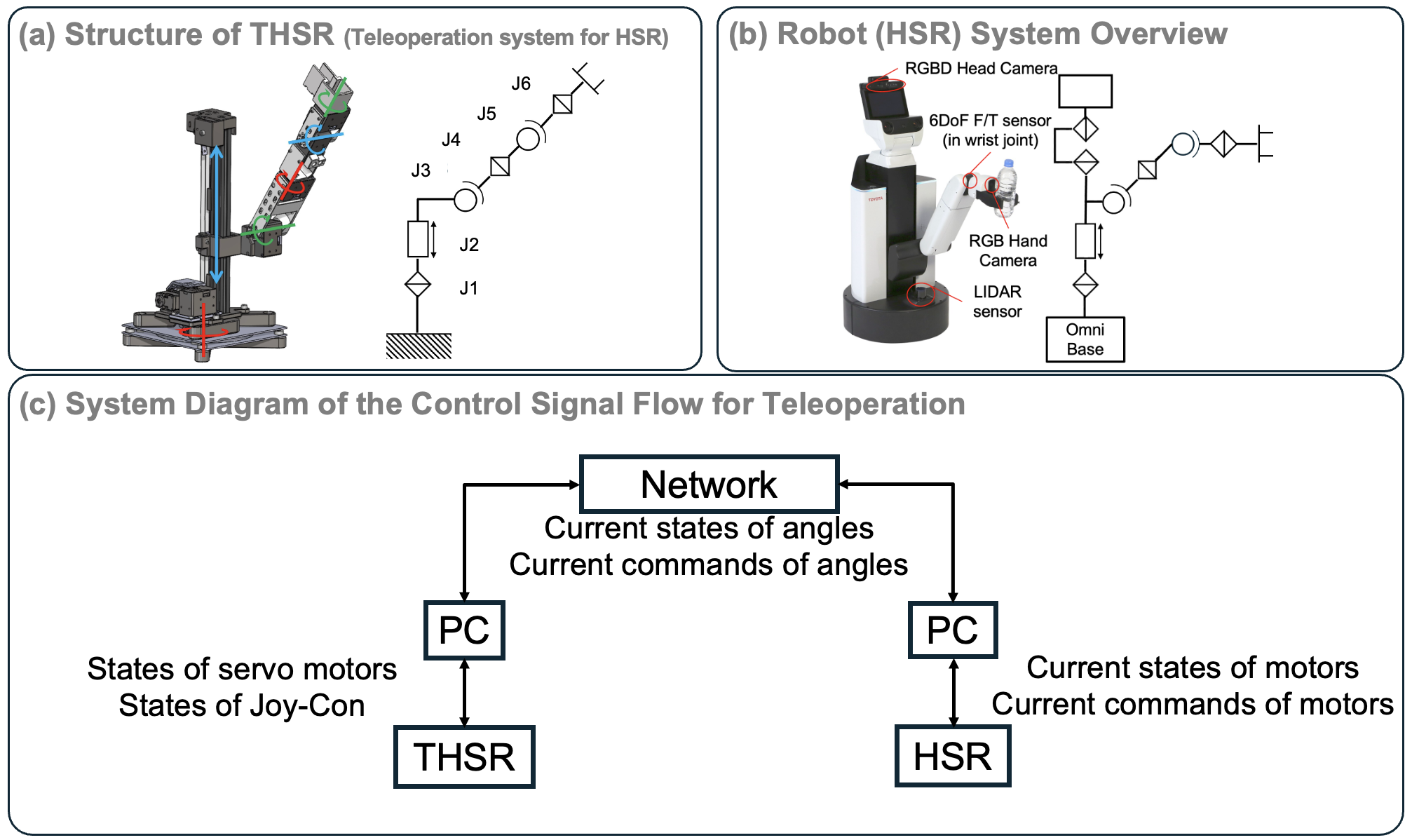}
  \caption{\textbf{The leader-follower teleoperation system used for data collection.} This system consists of \textbf{(a)} the leader device, THSR, manipulated by the operator; \textbf{(b)} the follower robot, HSR, equipped with multiple sensors; and \textbf{(c)} the control system that connects them. (a) The THSR's joints have a one-to-one mapping to the HSR's joints, enabling intuitive control without the need for inverse kinematics (IK) calculations. As shown in (c), inputs from the THSR's servo motors and an auxiliary Joy-Con controller are processed by a PC and sent over the network to the HSR, enabling the collection of high-quality manipulation data.
  }
  \label{fig:leader_follower_system}
\end{figure}

To address these challenges, we developed a one-to-one joint-mapping leader–follower device, THSR (Teleoperation system for HSR), which enables intuitive control of the HSR without requiring IK, as detailed below (see Fig. \ref{fig:leader_follower_system}).

This leader-follower system eliminates the need to solve inverse kinematics (IK), enabling intuitive control by ensuring the follower robot maintaining identical posture to the leader device. The THSR controls the 6-DOF position and orientation of the end-effector by transmitting joint commands from leader device joint states to the robot 4 joints along with the lifting axis and the base rotation axis. Auxiliary operations, such as mobile base translation, head camera orientation, and gripper actuation, are managed separately using a Joy-Con controller.

The hardware for each joint of the leader device employs Dynamixel servo motors serving for the joint angle measurements along with partial gravity compensation. The device is built using off-the-shelf mechanical components combined with parts fabricated with a standard FDM 3D printer. Key mechanical features include a hollow shaft for the base rotation axis, achieved through a belt-drive mechanism, which facilitates wiring and allows continuous rotation. For the lifting axis, a transmission mechanism using a timing belt is combined with a constant force spring to provide physical gravity compensation, enabling the device to hold its posture at any height.

The software control is implemented by directly sending the joint angles from the THSR's manipulator to the corresponding joints of the HSR as angle commands. For the lifting axis, the acquired joint angle is linearly transformed into a vertical displacement value and then scaled to fit within the HSR's operational range. The translational movement of the mobile base is controlled by velocity commands from the Joy-Con's joystick. For rotational movement, we implemented two modes: a constant-speed mode activated by button inputs and a proportional control mode based on the angular deviation between the THSR and the HSR. In addition to physical compensation, software-based gravity compensation is applied to the arm flex and wrist flex joints (see Fig. \ref{fig:leader_follower_system}).

Prior to collection, all 18 human operators underwent a training phase to ensure proficiency with the teleoperation system and the assigned tasks. Using this leader-follower teleoperation system, we collected data from a total of 18 human operators. The operators controlled the HSR to perform various household tasks. Control inputs for the base, arm, gripper, and head, along with data from various sensors, were recorded in the ROSbag format (a standard data logging format in ROS). Each episode was annotated with a success or failure flag, and the causes of failure—such as grasping errors, misalignment, or incorrect contact—were also documented. The actual dataset collection was carried out following the pipeline illustrated in Fig. \ref{fig:pipe}.

\begin{figure}
    \centering
    \includegraphics[width=1.0\linewidth]{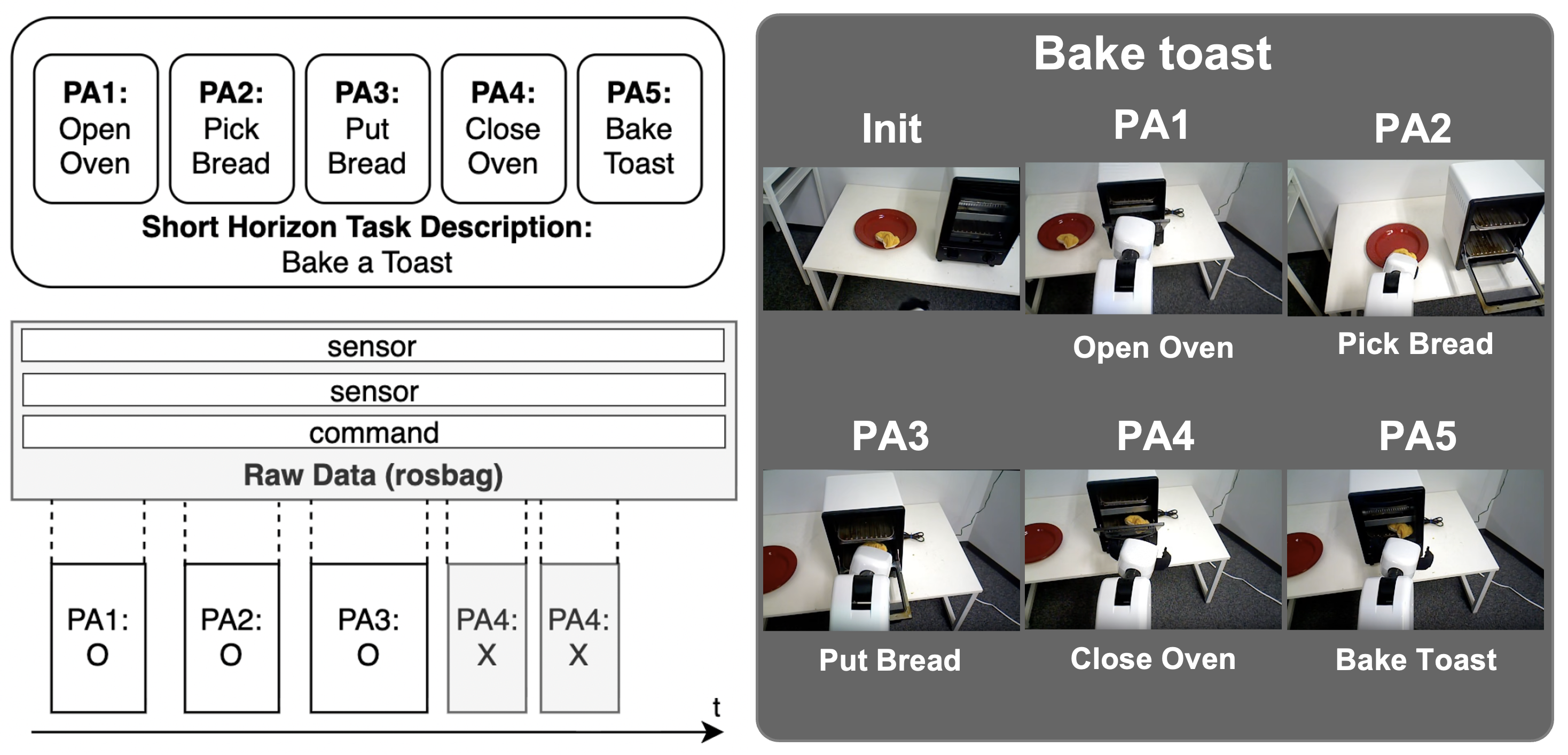}
    \caption{\textbf{An example of the hierarchical task annotation.} This figure illustrates how a high-level task, or Short Horizon Task, such as "Bake a Toast," is decomposed into a sequence of five Primitive Actions (PAs). Each PA (PA1: Open Oven, PA2: Pick Bread, PA3: Put Bread, PA4: Close Oven, and PA5: Bake Toast) corresponds to a specific segment of the time-series raw data (rosbag). Furthermore, each PA segment is annotated with a success (O) or failure (X) label, enabling detailed analysis of where errors occur within the overall task execution. 
    }
    \label{fig:task_hierarchy}
\end{figure}

\subsection{Task Hierarchy}
A core feature of our dataset is its two-layer annotation schema, designed to facilitate hierarchy decomposition.
\begin{itemize}
    \item Short Horizon Task (SHT): A high-level natural language instruction that defines the overall goal, e.g., ``Bake a toast.''
    \item Primitive Action (PA): A low-level command representing an atomic, semantically coherent segment. e.g., ``Open open,'' `Pick bread,` etc.. A sequence of PAs constitutes the completion of a SHT.
\end{itemize}

This structure is illustrated by the "Bake a Toast" task, which is decomposed into a sequence of five distinct PAs: 
PA1: Open Oven, PA2: Pick Bread, PA3: Put Bread, PA4: Close Oven, and PA5: Bake Toast. This hierarchical design enables models to learn both high-level task planning (the sequence of PAs) and low-level, sensorimotor control (the policy within each PA). The distribution of PAs and task durations across the dataset are shown in Fig. \ref{fig:task_hierarchy}.

\subsection{Data Quality \& Statistics}
To foster the development of robust and resilient policies, our data collection process explicitly included the recording and annotation of failure cases, which constitute approximately 6.6\% of the dataset. Such episodes may include grasping failure, manipulation error, problems during task execution, and they are marked during the process of data collection in the metadata. Additionally, we track the robot id, operator, and software versions used at different stages of data collection. This provides a valuable resource for research in error detection, recovery, and learning from negative examples.

To provide a deeper understanding of our dataset's composition, we present a statistical analysis in Fig. \ref{fig:dataset_statistics}. This analysis focuses on two key aspects: the distribution of fundamental skills and the duration of the tasks.
As illustrated in Fig. \ref{fig:dataset_statistics}(a), the dataset is heavily centered on core manipulation primitives. The most frequent skills are foundational actions such as "Grab," "Open," and "Place," which together constitute a significant majority of the collected trajectories. This distribution follows a distinct long-tail pattern, indicating a high density of data for common interactions, while more specialized skills are represented less frequently.
Fig. \ref{fig:dataset_statistics}(b) details the task duration distribution across all trajectories. The data reveals a strong concentration of tasks within the short-to-medium time frame, with the highest proportion of tasks being completed in 4 to 12 seconds. This characteristic suggests that our dataset is primarily composed of discrete, short-horizon activities. Consequently, it is particularly well-suited for training policies on fundamental, reactive behaviors, providing a robust foundation for learning essential robotic skills.

\begin{table}[t]
\centering
\small
\caption{\textbf{Breakdown of episodes and recorded hours per task in the dataset.} This table summarizes the total number of episodes, the cumulative duration of successful episodes, and the cumulative duration of failed episodes for each of the seven main Short Horizon Tasks. The inclusion of failure data is intended to support research into more robust policies, such as error detection and recovery. 
}
\label{tab:episode-summary}
    \begin{tabular}{rrrr}
        \toprule
        Short Horizon Task & \#Episodes & Successful Hours & Failed Hours \\
        \midrule
        Open the towel stand and hang the towel & 1315 & 18.65h & 0.49h \\
        Press the button to turn the desk lamp on and off & 746  & 1.57h  & 0.08h \\
        Bake a toast & 537  & 17.57h & 0.50h \\
        Make coffee & 483  & 38.56h & 4.27h \\
        Pull the chain to turn the desk lamp on or off & 401  & 5.50h  & 0.33h \\
        Stand the slippers in the slipper rack & 383  & 15.08h & 1.02h \\
        Washing dishes in the dishwasher & 336  & 47.50h & 2.96h \\
        \bottomrule
    \end{tabular}
\end{table}

\section{Supporting Frameworks: Processing and Standardization}

\subsection{Processing and Standardization}
\paragraph{Synchronization}
Since the sensors operated at different sampling frequencies, synchronization was required. 
Specifically, RGB images at 30 Hz, joint angles at 100 Hz, and force/torque signals at 100 Hz were all resampled at 30 Hz. In case of missing sensor samples we explicitly mark the sensor data for that frame as stale.
According to Lerobot v2.1 implementation, we, first, save images to disk in PNG format and then convert them to video files.
Episode boundaries are determined based on the multi-level task labels with recorded start and end timestamps.

\paragraph{Hierarchical Annotation}
Each episode was annotated with a two-layer structure. 
The higher layer corresponds to high level operation instructions, describing final goal with natural language 
(e.g., cleaning up the room, baking a toast, making a coffee). 
The lower layer corresponds to primitive actions, representing basic primitive operations such as grasping, moving, releasing, and opening/closing. 
The two layers form a nested structure, where a sequence of primitive actions corresponds to the achievement of a higher level goal.
This design enables hierarchical learning by separating high-level (high level instruction) and low-level (primitive actions) representations, and each layer having it's own success/failure labels allows flexible error analysis across different levels of task execution.

\paragraph{Standardization}
We utilze Hugging Face LeRobot v2.1 dataset format, which is widely adopted in open-source robotic community.
The use of standardization ensures direct compatibility with existing imitation learning methods and foundation models 
(e.g., RT-1, $\pi_0$, OpenVLA), thereby enabling reproducible cross-comparisons.

\subsection{Dataset Filtering and Packaging}
As depicted in the pipeline (Fig. \ref{fig:pipe}), once an episode is recorded, it undergoes an initial validity check to ensure all data streams were captured correctly. Valid episodes are then uploaded to a cloud storage for processing. The dataset preparation process involves both filtering and packaging, carried out in multiple stages. Initially, each individual ROSBag file is converted into a single LeRobot dataset, serving as an intermediate storage format. Once the initial filtering has been applied, these datasets are combined into a unified large-scale LeRobot package for the final release.

Filtering proceeds through three main stages: metadata queries, pre-processing, and post-processing. After data collection, we compile the dataset based on queries over metadata fields such as collection location, robot ID, and operator. We also conduct statistical analysis to detect outliers based on the distribution of episode durations.

During the pre-processing stage, we discard frame samples that do not contain even minimal motion. In the post-processing stage, we perform a final sanity check to ensure that episode data is intact and that all sensor values fall within valid ranges. Furthermore, we conduct statistical analysis of segment durations at the per-primitive action level, which allows us to identify and remove most outliers automatically. As a final step, we apply manual verification on randomly sampled episodes to detect any additional classes of outliers.
Once fully packaged and verified, the final dataset is prepared for data delivery to the research community through our public release.

\subsection{Privacy Considerations}
To prepare the first public release of the dataset, we evaluated several approaches for identifying episodes containing human appearances in data captured from both the head and hand cameras. Specifically, we compared a VLM-based approach using NVIDIA Cosmos-Curator with custom person-detection prompts against a conventional YOLO-based detector. The comparison revealed that having numeric metrics is advantageous, as thresholds can be tuned to reduce the number of false positives. Based on this observation, we adopted the YOLO-based detector, and episodes were automatically excluded whenever the number of detected frames exceeded a predefined threshold.

\section{Conclusion}

In this paper, we introduced the AIROA MoMa Dataset, a large-scale, multimodal dataset designed to advance mobile manipulation for long-horizon daily household tasks. The dataset captures over 90 hours of data spanning seven diverse household tasks, with rich observations including dual-view RGB, proprioception, and synchronized wrist force-torque signals. A novel two-layer hierarchical annotation of sub-goals and primitive actions provides structured supervision for hierarchical learning and detailed error analysis. The dataset is standardized in the LeRobot v2.1 format, enabling direct use with existing VLA models.

Beyond the dataset itself, we release an open-source pipeline for data conversion, filtering, and packaging, ensuring transparency and reproducibility. Looking forward, we plan to expand coverage by increasing the number of episodes, collecting from multiple sites and diverse environments, and incorporating recovery behaviors and human–robot interaction signals such as natural language and speech.

We believe the AIRoA MoMa Dataset establishes a critical benchmark for contact-rich, long-horizon mobile manipulation and will accelerate the development of the next generation of general-purpose robotic agents.

\section{Acknowledgements}
This AIRoA MoMa Dataset is based on results obtained from research conducted by AIRoA (AI Robot Association) with support from the ``GENIAC (Generative AI Accelerator Challenge)'' project, implemented by the Ministry of Economy, Trade and Industry (METI) and the New Energy and Industrial Technology Development Organization (NEDO), with the aim of strengthening Japan's development capabilities in generative AI.

\bibliographystyle{unsrt}
\bibliography{refs}

\end{document}